# Edge-Based Recognition of Novel Objects for Robotic Grasping

Amirhossein Jabalameli, Nabil Ettehadi, Aman Behal

***Abstract*—In this paper, we investigate the problem of grasping novel objects in unstructured environments. To address this problem, consideration of the object geometry, reachability and force closure analysis are required. We propose a framework for grasping unknown objects by localizing contact regions on the contours formed by a set of depth edges in a single view 2D depth image. According to the edge geometric features obtained from analyzing the data of the depth map, the contact regions are determined. Finally, We validate the performance of the approach by applying it to the scenes with both single and multiple objects, using a Baxter manipulator.**

## I. Introduction

A crucial problem in robotics is interacting with known or novel objects in an unstructured environments. Among the emerging applications, assistive robotic manipulators seek approaches to assist users to perform a desired object motion in an partial or fully autonomous system. While a wide research area is required to address this problem, our goal is recognizing a method that employs the robot visual perception to identify and execute an appropriate grasp to pick and place novel objects.

Finding a grasp configuration relevant to a specific task has been an active topic in robotics for the past three decades. In the recent review of Bohg et. al [31], grasp synthesis algorithms are categorized into two main groups, Analytical and Data-Driven. Analytical approaches explore for solutions through kinematics and dynamics formulations [23]. Object and/or robotic hand models are used in [8], [7], [3], [5] and [1] to develop grasping criteria such as force-closure, stable, dexterous and in equilibrium and evaluate if a grasp is satisfying them. The hardness in modeling a task, high computational costs and assumptions of the availability of geometric or physical models to the robot are the challenges that Analytical approaches deal with [23] in the real world experiments. Furthermore, researchers conduct experiments and infer the classic metrics are not sufficient to tackle with the grasping problems in the real world scenarios despite their efficiency in the simulation environments [18],[22].

On the other side, Data-Driven methods retrieve grasps according to their prior knowledge of either the target object, human experience or obtained information through acquired data. In compatible with this definition, Bohg et al [31] classified Data-Driven approaches depending on the encountered object is considered known, familiar or unknown to the method. Thus, the main point is how the query object is recognized and then compared with or evaluated by the algorithm existing knowledge. [10],[12] and [20] assume all the utilized objects are known for them, thus model object's shape with primitives such as boxes, cylinders and cones and define grasping strategy for each shape in the offline phase. Ultimately, match 3D mesh of the object in obtained data with their grasp database during the online phase.[15] exploits a probabilistic framework to estimate a pose of known object in an unknown scene. Ciocarlie et al. entered the human operator in the grasp planning control loop and mapped empirically to robot hand kinematics [16]. A group of methods, consider the encountered object as a familiar object and employ 2D and/or 3D object features to measure the similarities in shape or texture properties [31]. [13] trains a logistic regression model based on the labeled data sets and then detect grasping points for the query object depends on the extracted feature vector from a 2D image. Authors in [21], present a model that map the grasp pose to a success probability. The robot learn the probabilistic model through a set of grasp and drop actions.

The last group of methods in Data-Driven approaches, introduce and examine features and heuristics which directly map the acquired data to a set of candidate grasps [31]. They assume sensory data provide either full or partial information of the scene. [34] takes the point cloud and cluster it to a background and an object, then address a grasp based on principal axis of the object. Authors in [32] propose an approach that takes 3D point cloud and hand geometric parameters as the input. Then search for grasp configuration within a lower dimensional space satisfying defined geometric necessary condition. Jain et al. [33] analyze the surface of every observed point cloud cluster and automatically fit spherical, cylindrical or box-like shape primitive to them. The method utilizes pre defined strategy to grasp each shape primitive. The proposed algorithm in [28] build a virtual elastic surface by moving the camera around the object and compute the grasp configuration in an iterative process. [36] also approach the grasping problem through a surface-based exploration. Another approach to grasp planning problem can be performed through object segmentation algorithms to find surface patches [36], [29].

In general, knowledge level of the object, accessibility to partial or full shape information of the scene and type of the employed features are the main aspects that characterize Data-Driven methods. One of the main challenges that most of the grasping methods deal with and causes failure in real world experiments is robustness of a grasp against uncertainties in the sensed data and the execution phase. While [17] utilizes tactile feedback when performing the grasp to adjust the object position deviation from the initial expectation, [27] employs visual servoing techniques to facilitate the grasping execution. Another challenge in the object localization and grasping is background elimination which forces some of the method to make simplifying assumption about the objects situation such as [34] is validated for objects standing on a planar surface.

In this paper, we introduce a novel approach to generate proper hand configuration to grasp novel objects in an occluded scene using the objects partial shape information. Therefore, we develop a framework to localize the object visible edges instead of object detection. Using the extracted edges, we form contours representing the boundaries of the objects' surfaces. Then by evaluating all the contours based on geometric features of their belonging edges and end-effector constraints, we capture existed grasps. The obtained grasp satisfies force-closure, reachability and hand geometric specification. In addition, the grasp is characterized by introducing contact regions instead of contact points making it robust to uncertainties. The key idea of our method is to detect appropriate object edges for grasp extraction by processing a single view 2D depth image. Although a single 2D image does not take full advantage of existed information, it reduces the computational cost. Hence, the main challenge here is inadequate 3D information specially when there is an occlusion between objects.

This paper is organized as follows. Grasping preliminaries are presented in section II. The grasp problem is defined in section III. The proposed approach is presented in section IV, specifically in section IV.A., we define the objects model in the 2D image according to their geometries and then introduce the employed grasp model in section IV.B. Next in section IV.C., we propose an approach to find reliable contact regions for the force closure grasp on the targeted object. Details of implementing our algorithm are explained in section V. In section VI, we validate our proposed approach by considering different scenarios for placing objects, using a Kinect sensor and Baxter robot as a 7-DOF arm manipulator and discuss the obtained results. Finally, section VII concludes the paper.

## II. Preliminaries

Choosing a stable grasp is one of the key components of a given object manipulation task. According to the adopted terminology from [7], a stable grasp is defined as a grasp having force closure on the object. Force closure needs the grasp to be disturbance resistance meaning any possible motion of the object is resisted by the contact forces [30]. Thus, determining possible range of force directions and contact locations for robotic fingers is an important part of grasp planning [7]. By considering force closure as a necessary condition, [1] discussed the problem of synthesizing *planar grasps*. In the planar grasp, all the applied forces will lie in the plane of the object and shape of the object will be the only input through the process. Any contact between fingertips and the object can be described as a convex sum of three primitive contacts.

*Definition 1:* A wrench convex represents the range of force directions that can be exerted on the object and is determined depending on the contact type and the existing friction coefficient.

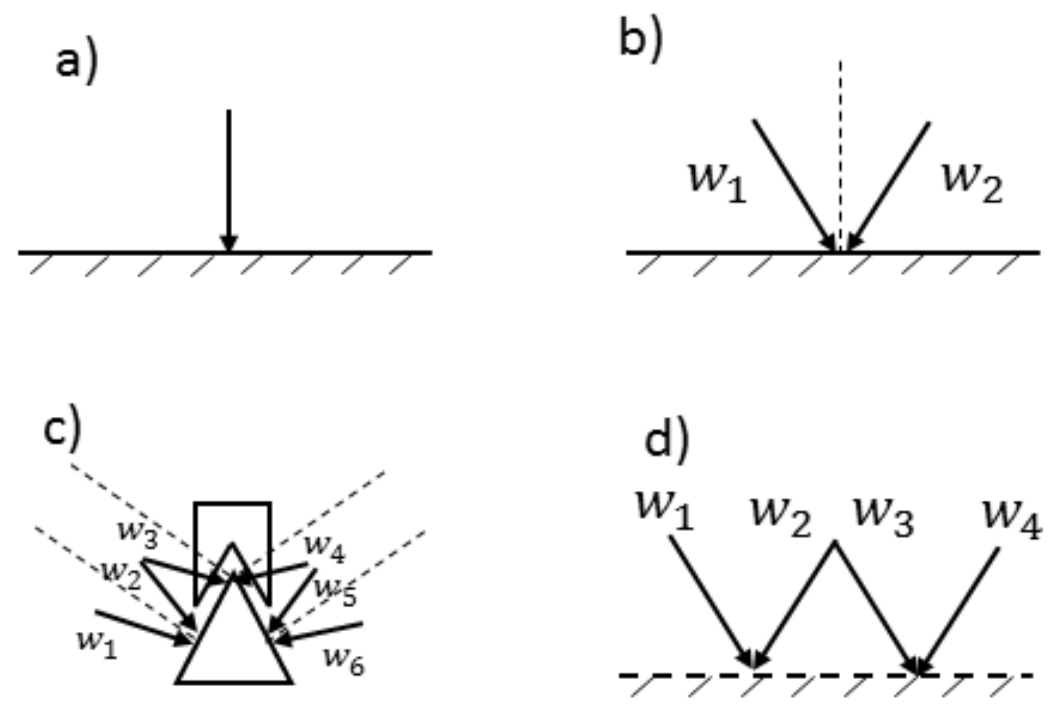

Fig. 1. Planar contacts: a) Frictionless point contact b) Point contact with friction c) Soft finger contact d) Edge contact

Figure (1) shows the primitive contacts and their wrench convexes in 2D. Wrench convexes are illustrated by two wrenches forming the angular sector.

In the frictionless point contact, the finger can only apply force in the direction of normal. However, through the point contact with friction, the finger can apply any forces pointing into the wrench convex. Soft finger contact is capable of exerting pure torques in addition to pure forces inside the wrench convex.

*Remark 2:* Any force distribution along an edge contact can be cast to a unique force at some point inside the segment. This force is described by the positive combination of two wrench convexes at the two ends of the contact edge.

It is also common that refer to convex wrench as friction cone in this subject. To resist translation and rotation motions for an 2D object, force closure is simplified to maintain force-direction closure and torque-closure [1].

*Theorem 3:* (Nguyen I) A set of planar wrenches $W$ can generate force in any direction if and only if there exists a set of three wrenches ($w_1, w_2, w_3$) whose respective force directions $f_1, f_2, f_3$ satisfy:
i) two of the three directions $f_1, f_2, f_3$ are independent.
ii) a strictly positive combinations of the three directions are zero: $\sum_{i=1}^{3} \alpha_i f_i = 0$

*Theorem 4:* (Nguyen II) A set of planar forces $W$ can generate clockwise and counter-clockwise torques if and only if there exists a set of four forces ($w_1, w_2, w_3, w_4$) such that:
i) three of four forces have lines of action that do not intersect at a common point or at infinity.
ii) let $p_{12}$(resp. $p_{34}$) be the points where the lines of action of $w_1$ and $w_2$ (resp. $w_3$ and $w_4$) intersect. There exist positive values of $\alpha_i$ such that $p_{34} - p_{12} = \pm(\alpha_1 f_1 + \alpha_2 f_2) = \mp(\alpha_3 f_3 + \alpha_4 f_4)$.

Basically, force-direction closure checks if the contact forces (friction cones) span all the directions in the plane. Torque closure tests if the combination of all applied forces produces pure torques. According to Theorem I and II, existence of four wrenches with three being independent is necessary for a force closure grasp in a plane. Assuming the contacts are with friction, each point contact provides two wrenches. Thus, a planar force closure grasp is possible with at least two contacts with friction. As stated by [3] and [1], conditions for forming a planar force closure grasp with two and three points interpreted in geometric as below and illustrated in figure (2):

- Two opposing fingers: A grasp by two point contacts, $p_1$ and $p_2$ with friction is force closure if and only if the segment $p_1 - p_2$ points out of and into two friction cones respectively at $p_1$ and $p_2$. Mathematically speaking, assuming $\varphi_1$ and $\varphi_2$ are angular sectors of friction cones at $e_1$ and $e_2$, term $\arg(p_1 - p_2) \in \pm(\varphi_1 \cap -\varphi_2)$ is the necessary and sufficient condition for two point contacts with friction.
- Triangular grasp: A grasp by three point contacts, $p_1, p_2$ and $p_3$ with friction is force closure if there exists a point, $p_f$ (force focus point) such that:
  i) for each $p_i$, the segment $p_f - p_i$ points out of friction cone of the $i$th contact.
  ii) let $k_i$ be the unit vector of segment $p_f - p_i$ which points out the edge. A a strictly positive combinations of the three directions are zero: $\sum_{i=1}^{3} \alpha_i k_i = 0$.

An appropriate object representation and analysis on the shape of objects based on the accessed geometry information is the first step toward finding contact regions for a stable grasp. In the next section, we relate planar object representation to a proper grasp representation in order to obtain the existing grasps.

## III. Problem Definition

The problem is addressed in this paper is to find contacting regions for grasping an unknown object. The obtained grasp has to be reachable, force closure and also feasible under the specification of the given end-effector. Partial depth information of the object which is sensed by the RGBD camera is the only input through this process and the proposed approach assumes the manipulated objects have rigid and non-deformable shapes. We also use the objects with non-transparent and non-reflective surfaces since they are not sensible by the employed sensor technology.

## IV. Approach

In this section, first we present an object representation and investigate its geometric features based on the scene depth map; then a grasp model for the end effector is provided.

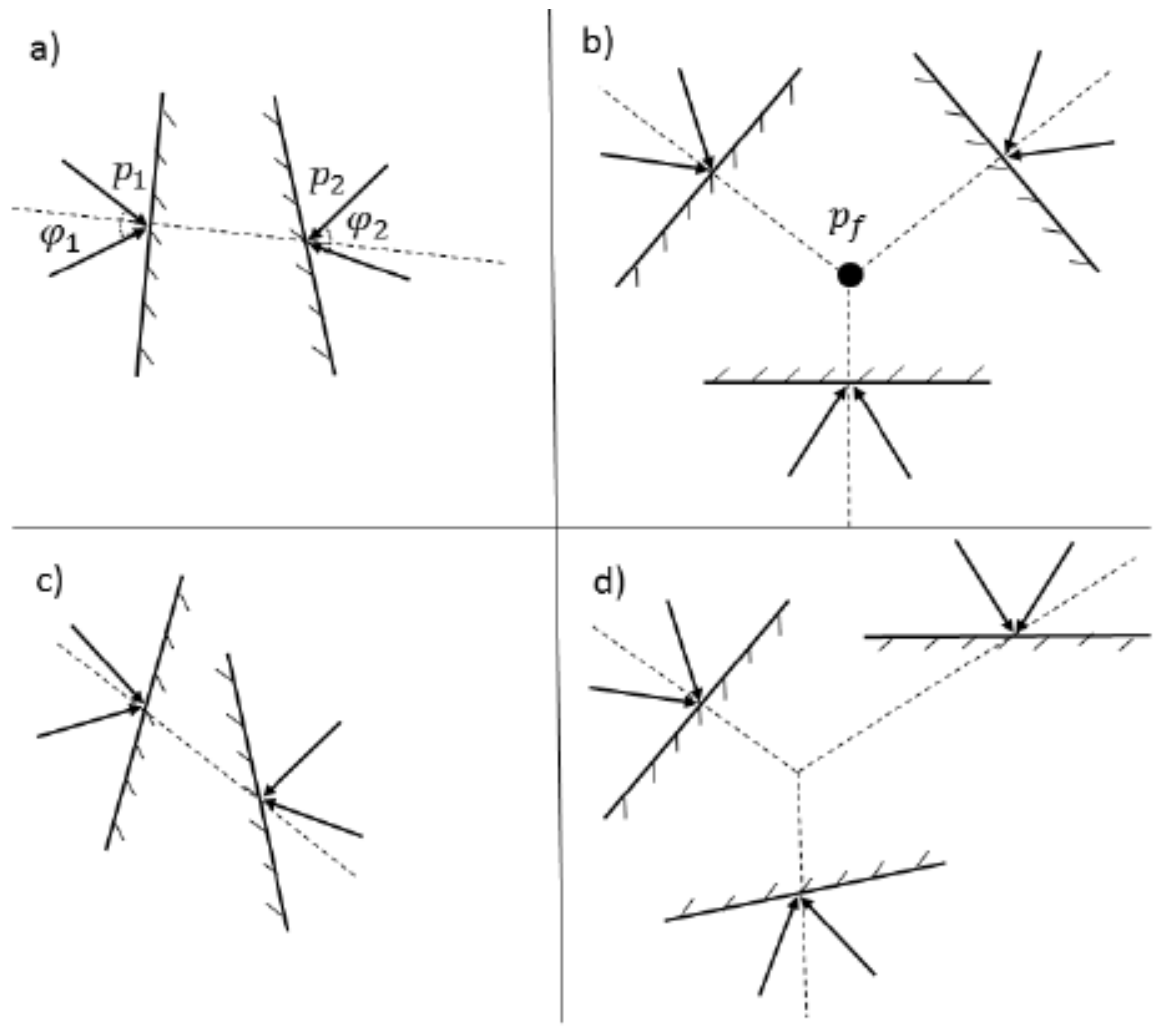

Fig. 2. Force closure geometric interpretation for two opposing finger gripper and triangular end effector. (a) and (b) show feasible force closures grasps, while (c) and (d) illustrate impossible force closure grasps.

In the end, pursuant to the development, we draw a relation between an object depth edges and force closure conditions. In addition, we specify contact location and end effector pose to grasp the target object.

### A. 2D Object Representation

Generally, 3D scanning approaches require multiple-view scans to construct complete object models. Due to physical limitations, we base our framework to represent objects by only utilizing partial information capturing from a single view. Consider an image of a cluttered scene including a variety of objects. We aim to partition the image into set of regions to represent meaningful areas, such that pixels (points) within a region share certain characteristics. Therefore, first we introduce elements of this representation.

*Definition 5:* A depth edge is a set of connected pixels in a 2D image where either distinct 3D surfaces appear to intersect in 2D or an object is partitioned from the background. $e_i = \{(u, v)|\text{cond. 1 or cond. 2}\}$

*Definition 6:* A closed contour is a set of depth edges which form a closed region. $C = \{e_i|make \text{ close region}\}$

*Definition 7:* A surface segment is a 2D region bounded by a closed contour. $S_i = \{\{u, v\}|\ f_i(u, v) = 0\}$

Hence, the visible part of the object in 2D is fully represented by either its depth edges, closed contours or surface segments.

$$O = \{\bigcup_i e_i\} = \{\bigcup_j Cj\} = \{\bigcup_k S_k\}$$

A surface segment is interpreted in 3D as visible part of a closed surface. Particularly, the surface segment fully characterizes corresponded 3D points for a plane. While in case of a curved surface it describes just a set of points available in the viewpoint. Mathematically speaking, we assume operator $\lambda : R^2 \rightarrow R^3$ maps 2D pixels to their real 3D coordinates. As a result, in figure (3) for the surface segment $S_1$ on the cube $\lambda(S_1) = A_1$, however, the surface segment on the cylinder $S_2$ implies only to a subset of cylinder's lateral surface in 3D $\lambda(S_2) \subseteq A_2$.

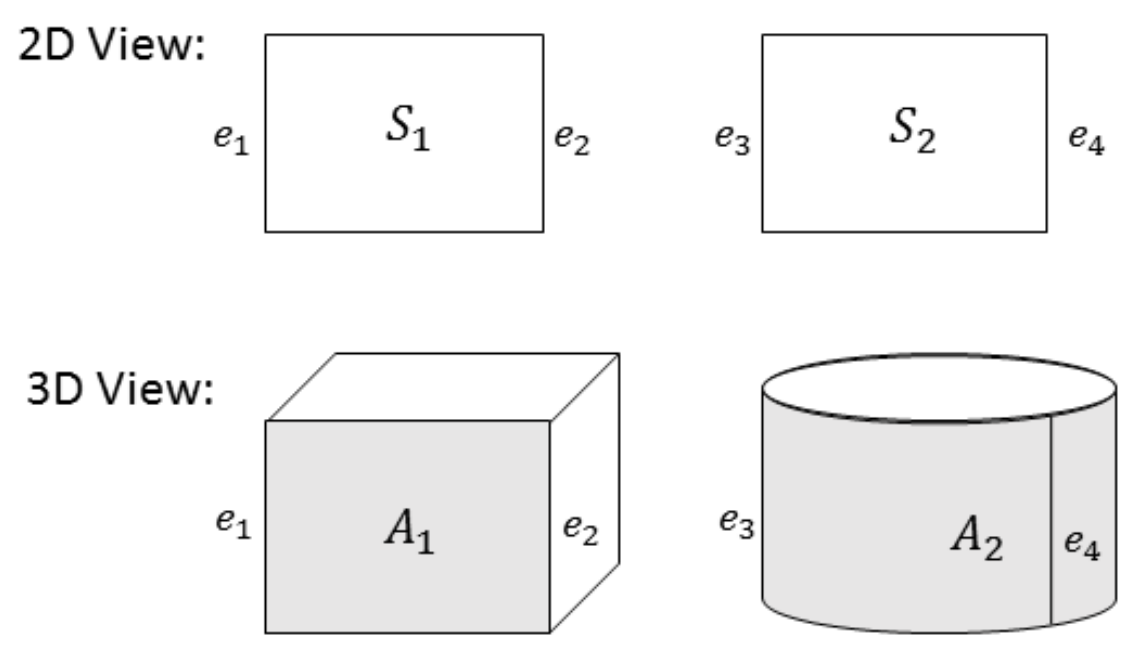

Fig. 3. Geometric interpretation of a surface segment for a cube and a cylinder.

Intuitively, the mapping of depth edges in 3D indicates intersection of distinct surfaces or the border of curved surfaces in the viewpoint. To expound kinds of depth edge and what they offer to the grasping problem we investigate their properties in depth map. All the depth edges are categorized into two main groups: 1. Depth Discontinuity (DD) edges and 2. Curvature Discontinuity(CD) edges. A DD edge is appeared by a significant depth value (distance to the camera) difference between its two sides in the 2D depth map and intimates a gap between its belonged surface and its surrounding along the edge. A CD edge is emerged by an abrupt change in the directional change of depth values. Although, it holds a continues change in depth values on its sides. Note that directional change of depth values is equivalent to surface orientation in 3D. In fact, a CD depth edge illustrates intersection of distinct surfaces in 3D.

A depth image can be described by 2D array of values which is described by an operator $d(.)$

$$z = d(r, c),\ d(.) : R^2 \rightarrow R$$

where $z$ denotes the depth value of pixel positioned at coordinates $(r, c)$ in the depth image $(I_d)$. Subsequently, gradient image, gradient magnitude image and gradient direction image are defined as follows

$$\begin{array}{lll} \text{Depth Image:} & I_d = & [d(r_i, c_i)] \\ \text{Image Gradient:} & \bigtriangledown I = & (\dfrac{\partial I_d}{\partial x}, \dfrac{\partial I_d}{\partial y})^T \\ \text{Gradient Magnitude Image:} & I_M = & [\sqrt{(\frac{\partial I_d}{\partial x})^2 + (\frac{\partial I_d}{\partial y})^2}] \\ \text{Gradient Direction Image:} & I_\theta = & [\tan^{-1}((\frac{\partial I_d}{\partial y})/(\frac{\partial I_d}{\partial x}))] \end{array} \tag{1}$$

where gradient magnitude image pixels describe the change in depth values in both horizontal and vertical directions. Similarly, each pixel of gradient direction image demonstrates the direction of largest depth value increase. In figure(7.c-d), color maps of depth image and gradient direction image is provided. Abrupt changes in the color intense is an indication of depth edges. Therefore, DD and CD edge pixels can be extracted from gradient magnitude image and gradient direction image, respectively. CD edges are also divided into two types, concave and convex. Consider a convex set $J$ encloses a CD edge in 2D image, the edge is called convex if function $d$ operates is convex or satisfy the following inequality:

$$\begin{array}{l} \forall j_1, j_2 \in J, \forall t \in [0, 1]: \\ D(tj_1 + (1-t)j_2) \leq tD(j_1) + (1-t)D(j_2) \end{array} \tag{2}$$

Otherwise, it is considered as a concave edge. Basically, outer surface of the object curves like interior of a circle at concave edges and curves like the circle's exterior at the convex edges.

*Remark 8:* Edge type determination extremely relies on the viewpoint.

*Remark 9:* A depth edge may or may not represent an actual edge of the object in real.

A concave CD edge holds its type in all the viewpoints. But a DD edge switches to convex CD edge or disappears (although indicating a convex) by changing the point of view. A convex CD edge may switch to DD edge by changing the point of view.

### B. Grasp and Contact Model

Generally, a precision grasp is addressed by end effector and fingertips poses with respect to a fixed coordinate system. According to the adopted terminology from [37], we refer to end effector $E$ with $n_E$ fingers and $n_\theta$ joints when fingertips contacting object's surface object, $O$, and define a grasp, $G$ as follows:

$$G = (\ p_G,\ \theta_G,\ C_G)$$

where $p_G$ is the end effector pose (position and orientation) relative to the object. $\theta_G = (\theta_1, \theta_2, ..., \theta_{n\theta})$ indicates the end-effector's joint configuration and $C_G = \{c_i \in S(O)\}_{i=1}^{n_E}$ determines a set of point contacts on the object's surface. The contact locations on the end effector's fingers is $C_E = \{\bar{c}_i \in S(E)\}_{i=1}^{n_E}$ and defined by a forward kinematics transform from the end-effector pose $p_G$.

Throughout this paper, we make an assumption regarding the end effector during the interaction with the object. Each fingertip applies force in the direction of its normal and the exerted forces by all fingertips will lie on a plane. We refer this plane and its normal direction as end effector's approach plane, $\rho_G$, and approach direction, $\vec{V}_G$. As a result, the contact points between the object and fingers will be located on this plane. In addition, some of the end effector geometric features can be described according to how they appear on the approach plane. For instance, grasp representation, fingers opening-closing range, finger's shape and width casted on a 2D plane is shown for a three finger end effector in figure (4).

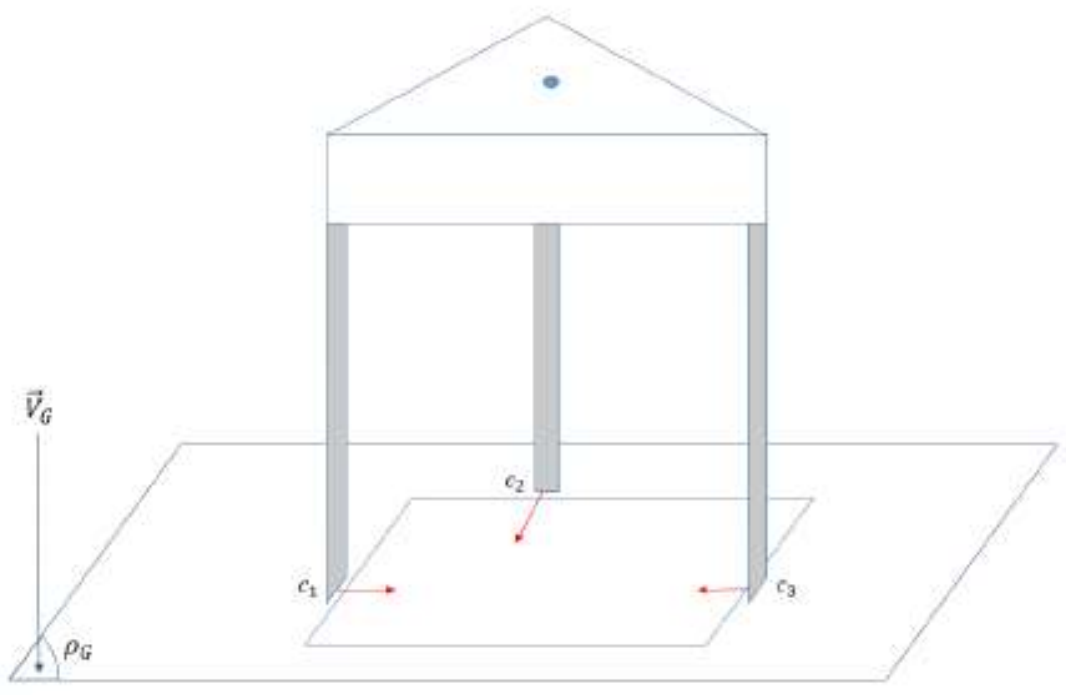

Fig. 4. Grasp representation for a planar shape

### C. Edge-Level Grasping

Experiments show human tendency to grasp the objects by contacting its edges and corners [1]. The main reason is, edges provide a larger wrench convex and accordingly a greater capability to apply necessary force and torque directions. To this point, we discussed how to extract depth edges and form closed contours based on available partial information. Each closed contour represents the object as a planar shape through a certain view. In this part, first we use the obtained contours as the input for planar grasp synthesis process. The output grasp will satisfy reachability, force closure and feasibility with respect to end effector geometric properties. Next, we analyze the conversion of a planar grasp to a executable 3D grasp. Finally, we point out the emerging ambiguity and uncertainties due to the 2D representation.

Reachability of a depth edge is measured by the availability of wrench convex lied in their belonged surface segment. While a convex CD edge provides two wrench convexes for its belonged surface segments, a concave CD edge is not accessible for a planar grasp. A DD edge is just reachable from one side. Therefore, DD and convex CD edges are remarked as reachable edges, while concave CD edges are not considered as the available points for the planar contact.

For the purpose of simplicity in the analysis and without loss of generality, we approximate curved edges by a set of line segments. As a result, all 2D contours turn into polygonal shapes. To obtain the planar grasp on a polygon, force closure construction mentioned in section II is reduced to evaluate all combinations of reachable edges subject to the following test:

- Angle test: Testing if the combination of the edge wrench convexes makes the force closure possible.
- Overlapping test: Checking the existence of a region on each edge providing contact locations subject to overlapping.

Output of angle test for a 2 opposing fingers is considered valid for a combination of edges if the angle made by two edges is less than twice the friction angle. The angle test for a 3 finger end effector is passed for a set of three contacts such that, a wrench from the first contact with opposite direction overlaps with any positive combination of the other two contacts provided wrenches. Overlapping test is also validated if there exists a contact region corresponds to each edge. To find overlapping area and corresponded region on edges taking the following steps are required:

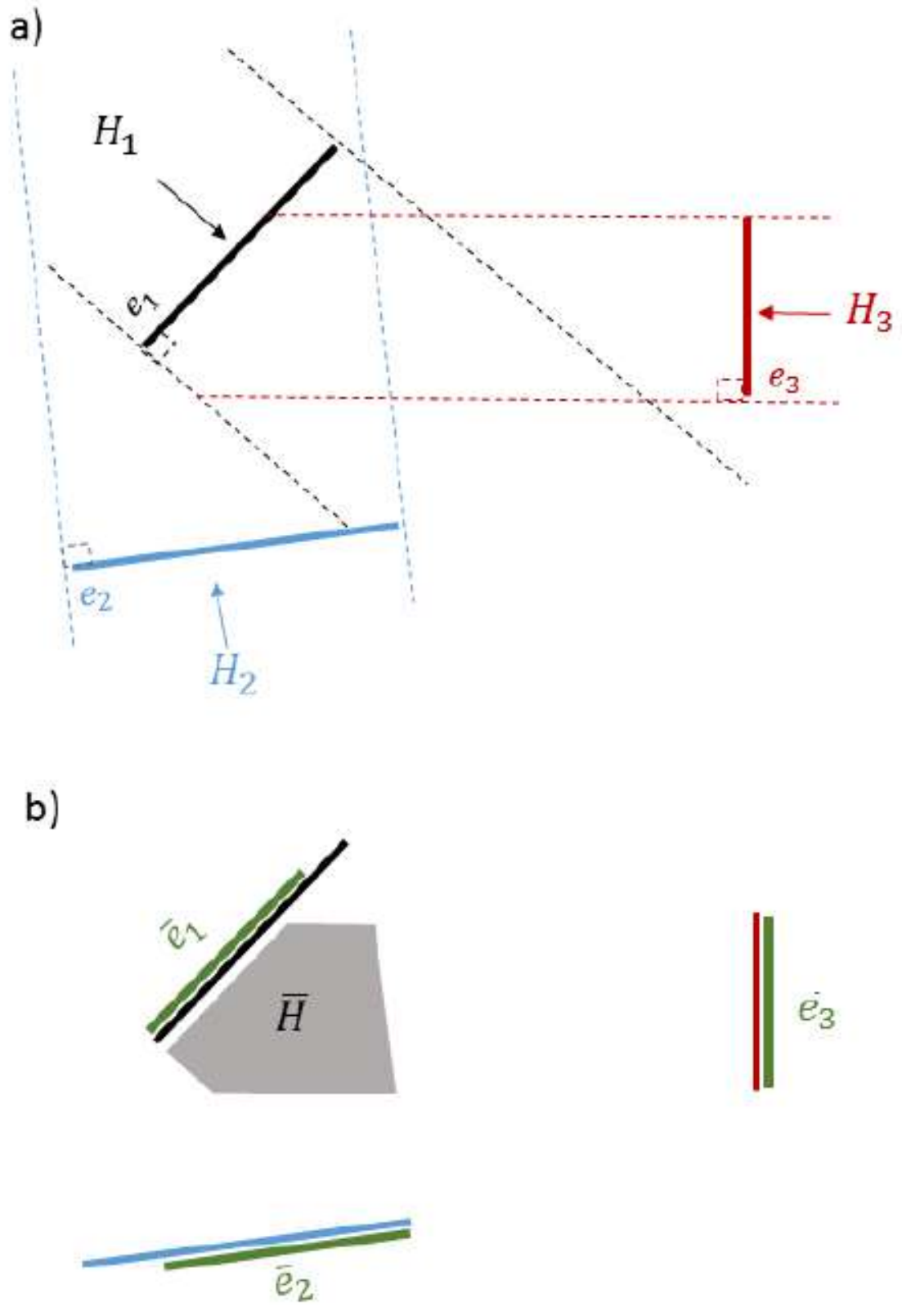

Fig. 5. Overlapping test. a) shows intersection of orthogonal projection for three edges b) indicates overlapped region and edge contact regions.

1) Form orthogonal projection area for each edge ($H_i$)
2) Find the intersection of projection areas by the candidate edges (Overlapping area $\bar{H}$)
3) Assigning the force focus point as the center of overlapping area ($p_f$)
4) Assigning the edge contact region by projecting overlapped region on each edge (edge contact region, $e_i^*$)

In fact, in this process, the planar force closure test is applied on the possible combination of reachable edges (polygon sides) with desired number of contacts belong to a certain contour. Figure (5) illustrates overlapping test for three edges.

To evaluate the feasibility of the output grasp with respect to the employed end effector, we move to the 3D space and extract 3D coordinates for the involved edges. Due to the assumption in section IV.B, all the applied forces through the fingertips lie on a plane. Hence, first we assure the existence of a 3D plane, includes all or a division of each edge contact region ($e_i^*$). This 3D plane determines end effector approach plane ($\rho_G$) and approach direction ($\vec{V}_G$) at the grasping moment. The geometric feasibility of the grasp is constrained by the end effector parameters, such that the fingers locate with the certain distance. For instance, a two-fingered gripper width range specifies if the end effector can fit around a pair of edges. Depending on the end effector, other parameters such as fingertips width can also be measured for outputing a valid grasp.

According to presented grasp model in section IV.B, we desire to determine contact points for each fingertips. In order to make the grasp robust to positioning errors, the center of edge contact regions ($e_i^*$) which lie on the fit plane denotes point contact on the object surface ($c_i$). We discuss specification of the end effector pose in the implementation section, since it depends on end effector kinematics and the hired grasp policy for execution.

To sum up the discussed approach we draw the steps in the following. First, we extract all the depth edges and form closed contours. In the second step, depth edges forming each contour are evaluated to satisfy reachability and planar force closure conditions. Next, we consider the existence of a plane using for each combination of edges and feasibility with respect to the end effector geometric properties. In the final step, the grasp parameters are determined based on the extracted plane and edge contact regions. The following algorithm also describes searching steps for constructing force closure using a depth image:

1) detect disc edges from depth image
2) detect curvature disc edges from gradient direction image
3) form closed contours using all depth edges
4) for each contour:
   a) remove the concave CD edges
   b) make combination of edges with desired number of contacts
   c) for each edges combination:
      i) perform angle test
      ii) perform overlapping test and output edge contact regions in 2D
      iii) perform plane existence test on 3D coordinates and output end effector approach plane and contact locations
      iv) perform end effector geometric constraint test
      v) output grasp parameters w.r.t end effector kinematics

As a matter of fact, each depth edge appearing in the 2D image is shared between two surface segments, which at least one of the surface segment is always visible in the view. In the planar force closure grasp, end effector exerts force to depth edges to manipulate just one of the two linked surfaces. Considering geometry of the connected surfaces is an important factor for analyzing the extension of planar force closure to 3D force closure. Based on the current layout, our approach provides similar grasp for the contact pairs $(e_1, e_2)$, $(e_3, e_4)$ and $(e_5, e_6)$ in figure(6). While the depth edges in these three cases offer different 3D wrench convexes which impact the grasping. Throughout this paper, we assume the existence friction between fingertips and object is large engough such that applied planar forces lie inside 3D friction cones at the contacting points. Another impacting factor is relative position of force focus point with respect to the gravity center of the object. Applying sufficient level of force prevents possible torques that ensued from this uncertainty.

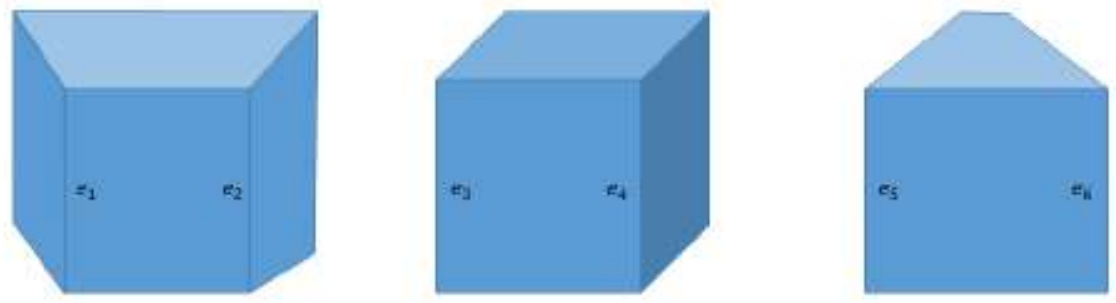

Fig. 6. Shapes with similar planar grasps despite different 3D friction cones.

## V. Implementation

In this section, we describe the implementation steps to process a depth image as the input and identify appropriate grasps. Notice that the current implementation focuses on finding grasps for a two-opposing finger gripper. Therefore, we employ the described algorithm in section IV.C to construct a grasp based on forming combination of two edges to indicate a pair of contact locations. A set of pixel-wise techniques is utilized to achieve the regions of interest in a 2D image and to address the desired 3D grasp. In addition, to cope with noise effects of edge detection step in the algorithm, we utilize a tweaked procedure to follow the approach steps. In fact, we skip contour formation process in the third step of the approach and directly look for the pairs that meet the discussed conditions. Thus, if an edge is missed in the detection step, we do not lose the whole contour and its corresponding edge pairs. However, the emerging complication is expansion of the pair formation search space. Later in this section, we introduce constrains to restrict this search space.

### A. Edge Detection and Line Segmentation

According to section IV.B, depth edges appear in depth image and gradient direction image. Due to the discontinuity existing by traveling in the orthogonal direction of a DD edge in depth image ($I_d$), the edge belonging pixels are local maximums of $I_M$ (magnitude of the gradient image $\bigtriangledown I$). Alongside, a CD edge demonstrates a discontinuity in gradient direction image ($I_\theta$) values, which illustrate a sudden change in normal directions corresponded to the edge neighborhood. Thus in the first step, an edge detection method is required to be applied to $I_d$ and $I_\theta$ to capture all the DD and CD edges, respectively. We selected Canny edge detection method [2] that outputs the most satisfying results with our collected data.

Generally the output of an edge detection method is a 2D binary image. Imperfect measurement in depth image yields to appearance of artifacts and distorted textures in the output binary images. For instance, an edge in the ideal way is marked out with one pixel-width. However, practically there exist non-uniform thickness along the detected edges In order to reduce such effects and enhance the output of the edge detection, a set of morphological operations are applied to the binary images. In coordination with the mentioned attempt, logical OR operation is used to integrate all the marked pixels corresponding to depth edges from $I_d$ and $I_\theta$ in a single binary image called detected depth image $I_{DE}$. Figure (7) shows the output of edge detection step for an acquired depth image from the Object Segmentation Dataset [25]. Note that, the only input through the whole algorithm is $I_d$, and color image is merely used to visualize the obtained results. For the purpose of visualizing, a range of colors are also assigned to the values of $I_d$ and $I_\theta$. Improvement made by the morphological operations is noticeable in Figure 7.d.

To perform further process, a procedure is required to distinguish edges by a 2D representation in the obtained binary image ($I_{DE}$). Considering a 2D image with the origin on the left bottom corner, each pixel is addressed by a pair of positive integers. We employed a method by [38], to cluster binary pixels into groups and then represent them by start and end points. Given $I_{DE}$, we first congregate the marked pixels into connected pixel arrays, such that each pixel in an array is connected only to pixel(s) in its 8 immediate neighbor of the same array. Next, an iterative line fitting algorithm is utilized to divide the pixel arrays into segments, such that each segment is indicated by its two end-points. The pixels belong to a segment, satisfy an allowable deviation threshold from the 2D line formed by the end-points. As a result, a straight edge corresponding pixels are represented by one line segment. While, curve edges are captured by a set of line segments. Figure(8) indicates outputs of marked edge pixels and corresponded line segmentation for a synthetic depth image; colors are randomly assigned to distinguish the captured lines. Operator $|L_i|$ computes pixel-length of line segment and $\measuredangle(L_i)$ measures the angle which is made by the line segment and the positive direction of horizontal axis. Assuming the line segment always points out and counter clockwise as the positive orientation, $\measuredangle(.)$ outputs an angle in the range of $[0^o \ \tilde{} + 180^o)$.

### B. *Edge Feature Extraction and Pair Formation*

By end of previous step, a set of pixel groups, indicated by a corresponded set of line segments, are provided. In this part, we aim to form pairs of line segments subject to mentioned constraints in section IV.C. We define local and mutual geometry features extracting from edge neighborhoods. Although, mathematical relations of features rely on single pixels. We create 2D masks enclosing the line segment. Consider operator

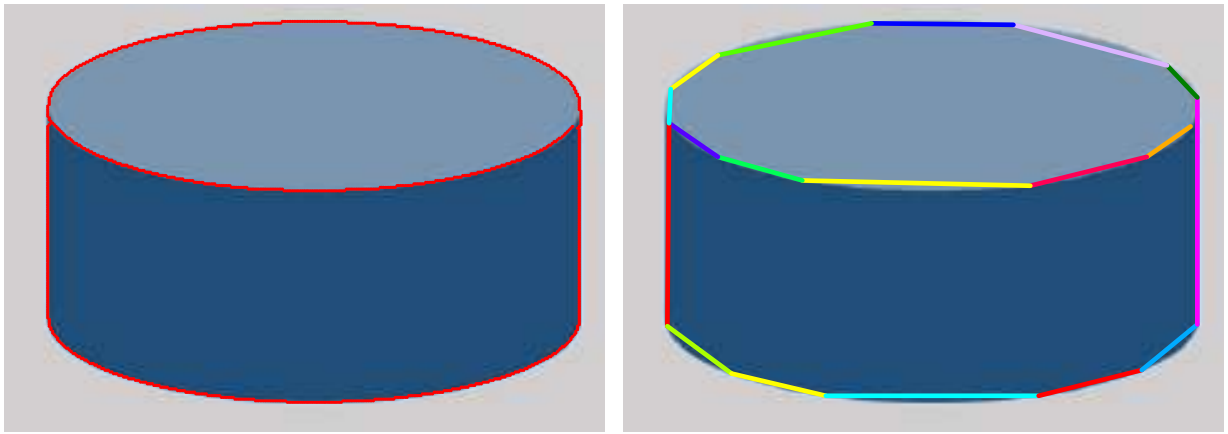

Fig. 8. Line segmentation step is applied to a synthetic depth map. (a) detected edge pixels are marked) (b) edges are broken into line segment(s)

$h(.)$ locates the region of interest. A parallelogram binary mask can be obtained by

$$h(\vec{L}_i, \vec{W}) \equiv h(\vec{L}_i, (w, \gamma))$$

where $\vec{L}_i$ and $\vec{W}$ are the sides. In the equivalent operator representation, $w$ shows pixel-length of the line segment $\vec{W}$, while $\gamma$ denotes the angle between sides $\vec{W}$ and $L_i$ in the range of $[-180^{o \, \tilde{}} + 180^o)$. In a similar way, we provide the following predefined masks for a line segment:

$$\begin{aligned} H^0(L_i) &= h(\vec{L}_i, (1, \ \ +90)) \\ H^+(L_i) &= h(\vec{L}_i, (w_0, \ \ +90)) \\ H^-(L_i) &= h(\vec{L}_i, (w_0, \ \ -90)) \end{aligned}$$

Applying kernels build upon $h(.)$ to depth image and the other constructed images help to make the feature identification process robust. Figure (9.a) demonstrates masks $H_1$ and $H_2$ provide a positive angle parallelogram for $L_1$ and negative angle parallelogram for $L_2$, respectively.

First, we evaluate reachability of each line segment and existence of a wrench convex for it. To do so, the line segments have to be assigned with edge type label. Comparison of binary masks $H^0(L_i)$ applied to $I_d$ and $I_\theta$ images, results to distinguish DD and CD line segments from one another. In addition, a line segment divides its local region to two sides. Therefore, the object is posed whether on the side with a positive orientation w.r.t. the line segment or a negative orientation. As discussed in section IV.B, the wrench convex(es) is available in certain side(s) for each line segment. Note that, depth value of DD edge sides hint at object relative pose with respect to the line segment. As a result, the side with lower depth value implies object (foreground), the side with greater depth value points out the background and correspondingly available wrench is suggested. Likewise, evaluating the sides and line segment average depth values based on equation (2), yields to specify convexity/concavity of a CD edge. Mathematically speaking, edge type feature is determined for a DD line segment $L_i$ and a CD line segment $L_j$ as follows :

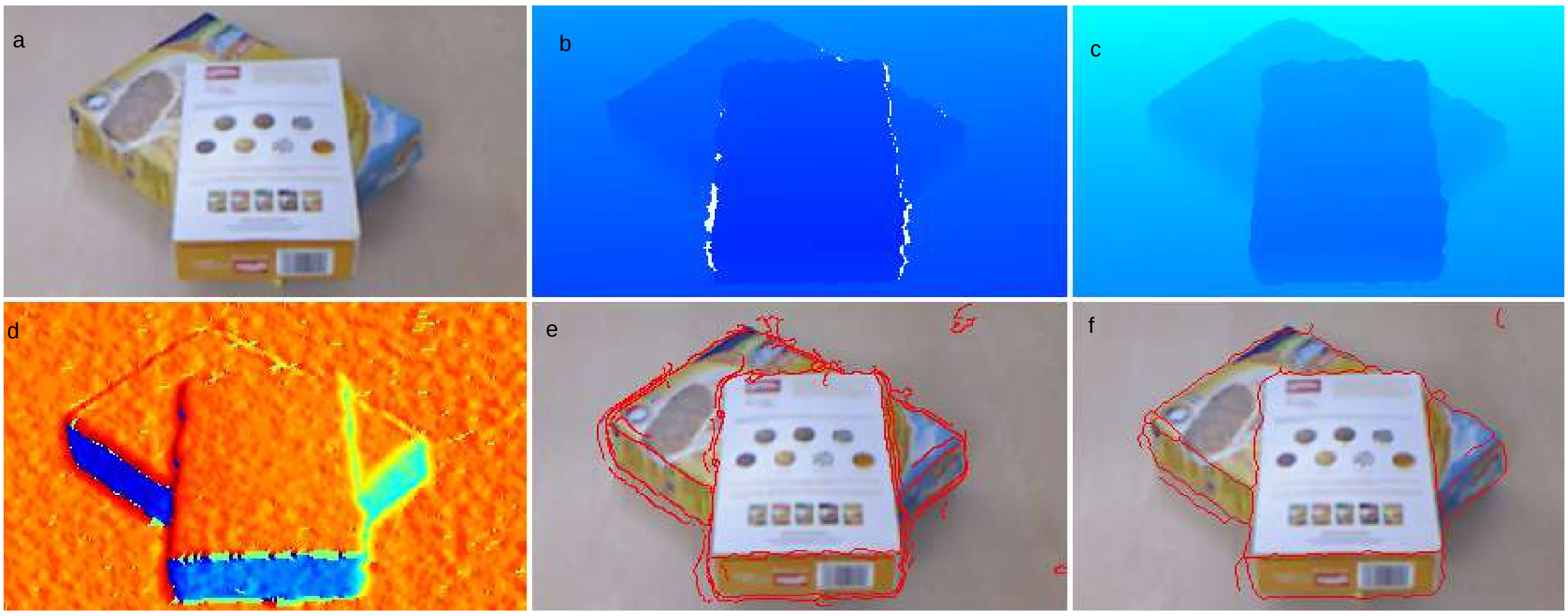

Fig. 7. Applied edge detection on an acquired depth map. (a) RGB image of the scene, $I_c$ (b) Color map of the raw depth map. White pixels imply to non-returned values from the sensor (depth shadows) (c) Color map of the processed depth map, $I_d$ (d) Color map of computed gradient direction image, $I_\theta$ (e) Detected edges before applying the morphological operations (f) Detected edges after the morphological process, $I_{DE}$.

$$\begin{cases} \text{if}: \bar{d}(H^+(L_i)) < \bar{d}(H^-(L_i)) \\ \text{then}: L_i \text{ is } DD^- \\ \text{otherwise}: L_i \text{ is } DD^+ \end{cases}$$

$$\begin{cases} \text{if}: 1/2[\bar{d}(H^-(L_j)) + \bar{d}(H^+(L_j))] > \bar{d}(H^0(L_j)) \\ \text{then}: L_j \text{ is } CD^\pm \\ \text{otherwise}: L_j \text{ is } CD^0 \end{cases}$$

such that $(\pm, +, -, 0)$ signs indicate availability of wrench convex w.r.t the line segment and $\bar{d}(.)$ operator takes average of depth values over the specified region.

The pair of $(L_i, L_j)$ represents a planar force closure grasp for 2-opposing finger, if line segments have opposite wrench signs and satisfy the following conditions obtained from section IV.C:

$$\begin{cases} |\measuredangle(L_i) - \measuredangle(L_j)| < 2\alpha_f \\ \bar{H}_\beta \neq \phi \end{cases}$$

where $\alpha_f$ is determined by the friction coefficient. The $\bar{H}_\beta$ mask is the pair overlapping area which is captured by intersection of edges projections and acquired by the following relations:

$$\bar{H}(\beta) = H_\beta(L_i) \cap H_\beta(L_j)$$
$$H_\beta(L_i) = \begin{cases} h(\vec{L}_i, (w_{\max}, \beta)) \text{ if } DD^- \text{ or } CD^- \\ h(\vec{L}_i, (w_{\max}, -\beta)) \text{ if } DD^+ \text{ or } CD^+ \end{cases}$$
$$\beta = 1/2 \times |180 - |\measuredangle(L_i) - \measuredangle(L_j)||$$

such that $H_\beta(L_i)$ addresses projection area made by line segment $L_i$ with the angle of $\beta$. In fact, $\beta$ implies orthogonal direction of the bisector. Assuming existence of the overlapping area, edge contact regions, $L_i^*$ and $L_j^*$ are parts of the line segments which enclosed by the $\bar{H}(\beta)$ mask. Figure (9.b) demonstrates projection areas and contact regions for a pair of edges.

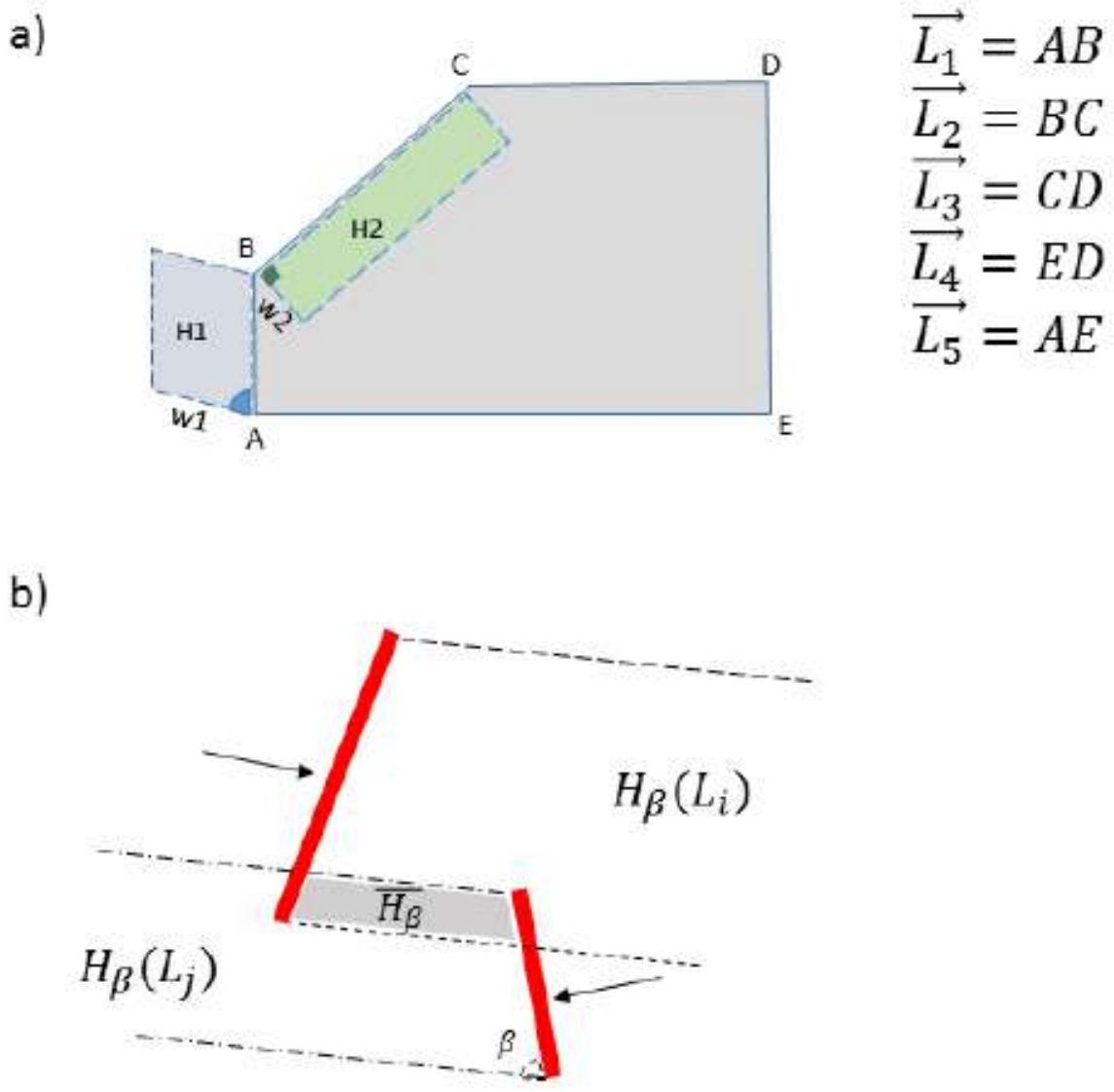

Fig. 9. a) Examples of parallelogram masks the sides of 2D shape ABCDE. b) Projection area and edge contact regions for a pair of edges.

*Remark 10:* In a case that we have access to the closed contours formed by depth edges, both the line segments are required to belong to a same closed contour.

To this point, planar reachability and force closure features are assessed. As the final step, we check if the pair is feasible under the employed gripper constraints. We assume $P_i = \lambda(L_i^*)$ is the set corresponding all the 3D points located on $L_i^*$ region. Euclidean distance between the average points of two sets $P_i$ and $P_j$ is required to satisfy:

$$\epsilon_{\min} < ||\bar{P}_i - \bar{P}_j||_2 < \epsilon_{\max}$$

where $\epsilon$ denotes the width range of the gripper. In addition, to assure that $P_i$ and $P_j$ posed on a plane, we fit plane model to the data. Throughout the current implementation, we utilized RANSAC method to estimate the plane parameters. The advantage of RANSAC is its ability to reject the outlier points emerged by the noise. If a point holds greater distance from the plane than an allowable threshold ($t_{\max}$), is considered as an outlier point. The output plane and the normal unit vector pointing in the plane are referred as $\rho_R$ and $\vec{V}_R$. Note that, for further processes, sets $P_i$ and$P_j$ are also replaced with corresponding sets excluding the outliers.

### *C. 3D Grasp Specification*

We desire to calculate grasp parameters based on the presented model in section IV.B. To reduce the uncertainties effects, we pick the centroid of the edge contact regions ($P_i$) as the safest contact points. As stated by [19], a key factor to improve the grasp quality is orthogonality of the end effector approach direction to the object surface. In addition, the fingers of a parallel-finger gripper can only move toward each other. Hence, according to the employed grasp policy, the gripper holds a certain pose such that the gripper approach direction is aligned with normal of the extracted plane. In the meantime, closing the fingers yields to contact the object at the desired contact points. Thus, for a graspable pair, grasp parameters are described by:

$$\begin{aligned} G(L_i, L_j) &= (p_G, \theta_G, C_G) \\ &= \begin{cases} p_G = (P_G, \mathbf{R}_G) \\ \theta_G = \{\theta_1, \theta_2\} \\ C_G = \{c_1, c_2\} = \{\bar{P}_i, \bar{P}_j\} \end{cases} \end{aligned}$$

where 3D vector $P_G$ and rotation matrix $\mathbf{R}_G$ indicate the gripper pose. We adjust $\theta_G$ such that fingers take maximum width before contacting and a width equals to $||\bar{P}_i - \bar{P}_j||_2$ during the contact. If length of the fingers are equal to $\epsilon_d$ and the fingers direction closure is defined by the unit vector $\vec{V}_c = (\bar{P}_i - \bar{P}_j) \ / \ ||\bar{P}_i - \bar{P}_j||_2$, then we can obtain:

$$\begin{cases} P_G = 1/2 \times (\bar{P}_i + \bar{P}_j) - \epsilon_d \vec{V}_G \\ \mathbf{R}_G = \mathbf{R}_{o_1}^{o_2} \\ \vec{V}_G = \vec{V}_R \end{cases}$$

The matrix $\mathbf{R}_{o_1}^{o_2}$ represents a rotation from the world coordinate frame $o_1$to the coordinate frame $o_2$ which is captured by three orthogonal axis $[\vec{V}_R; \vec{V}_c \times \vec{V}_G; \vec{V}_G]$.

### *D. Practical Issues*

Through the process of implementation on the real data, we face issues which are caused by uncertainties in the measured data. According to [24] error sources for imported data by depth sensors origin from imperfect camera calibration, lighting condition and properties of the object surface. RGBD sensors are subject to specific problems in measuring depth information based on the technology they use [35]. A common example of these problems is shadows or holes that appear in the depth image which point out the sensor inability to measure depth of such pixels. The main reason is some regions are visible to the emitter but not to the receiver sensor. Consequently, the sensor returns a non-value code for these regions. Since our implementation is mainly dominated by pixel-level processes, a procedure is required to handle this issue. In order to do so, we use a recursive median filter to estimate depth values for the shadow regions [26]. In Fig. **??**.b, white pixels display shadows in the sensed depth image and Figure **??**.c demonstrates depth image after the estimation procedure is performed.

Another issue in the case of DD edges is, if each edge pixel is rightly placed on its belonged surface. In practice, an edge is detected as a combination of the pixels placed on both the object and the background. Since the marked pixels are utilized for the purpose of object pose estimation, we are interested to locate them on the foreground object. Although there are efficient ways to recognize the foreground pixels such as [9], but due to the computational cost we make use of a very simple pixel-level procedure. Based on the object-line relative position, we create $H^+$ or $H^-$ mask that orients toward the object side. Applying the mask to the gradient magnitude image ($I_M$) provides accurate location of maximum depth gradient along the mask width (perpendicular to the line segment). The relative position of marked edge pixels with respect to the peak of the gradients determines if they are located on the object side or not. In the case of incorrect allocation, we move the marked pixel in the direction perpendicular to the line segment with a sufficient displacement to make sure the new marked pixel is located on the foreground side. It is important to note that this process is applied only to DD edges since there is no foreground/background concept for a CD edge.

Due to the projection occurring in camera from 3D to 2D, an ambiguity emerges causing two distinct depth edges along each other being captured as a single line segment. This issue can handled by adding extra examination to the edge feature extraction. Considering masks $H^0$ applied to $I_d$ and $I_\theta$ images demonstrate if there exist any depth or gradient orientation discontinuity along the edge. The occurrence of this discontinuity yields to breaking the edge into two line segments at the location of the abrupt change. Consider that in the existence of closed contour formation, the ambiguity does not arise and this test is discarded. In figure () edges $e_1$ and $e_2$ in 2D image are considered as one line segment, while by performing the above test, they can be distinguished.

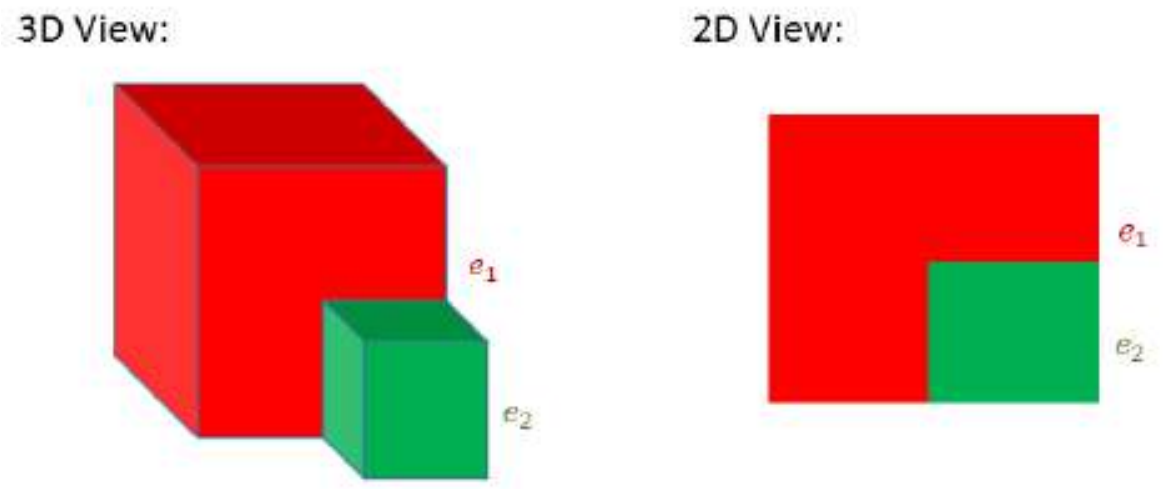

Fig. 10. 2D object representation ambiguity.

## VI. RESULTS

In this section, first we evaluate the performance of detection step of grasp planning algorithm and then conduct two experiment setups to test the overall grasping performance using the 7-DOF Baxter arm manipulator. A standard data set named Object Segmentation Database (OSD) [25] is adopted for the simulation. Besides, we collected our data set using Microsoft Kinect One sensor for real world experiments. The data sets include a variety of unknown objects from the aspects of shape, size and pose. In both cases, the objects are placed on a table inside the camera view and data set provides RGBD image. The depth image is fed in the grasp planning pipeline and RGB image is just used to visualize the obtained results. Note that all the computations are performed in MATLAB.

### A. Simulation-Based Results

In this part, to validate our method we focus on the output results of detection step in a simulation-based environment, *i.e.*, edge detection, line segmentation and pair evaluation. To do so, we chose 8 images from OSD dataset including different object shapes and cluttered scenes. Figure (11) shows provided scenes.

To specify the ground truth, we manually mark all the reachable edges (DD and Convex CD) for the existing objects and consider them as *graspable edges*. If each graspable edge is detected with correct features, it is counted as a *detected edge*. Assuming there is no gripper constraints, a *graspable surface segment* is determined if it provides at least one planar force closure grasp in the camera view. In the similar way, detected surface segment , graspable object and detected object are specified. Table (VI-A) shows the obtained results by applying the proposed approach on the data set. In addition, Figure (12) illustrates the ground truth and detected edges for scene No.4.

According to the provided results, although 20% of the graspable edges are missed in the detection steps, 97% of the existed objects are detected and represented by at least one of their graspable surface segments. This emphasizes how skipping the contour formation step has positive effects through the grasp planning. Obtained results also indicate the efficiency of the proposed approach decreases as the scene becomes more cluttered. Addressing how exactly the performance of these pixel-wise techniques, such as edge detection and morphology operations, affect the efficiency of the our approach is complex. Output quality and setting of these methods strongly depend on characteristics of the image view and scene. Therefore here, we only analyze edge length effects and avoid detailing other effective parameters.

In fact, an edge appearing longer in a 2D image is composed of a greater number of pixels. Thus, it has a slighter chance to be missed in the detection step. In addition, since there is uncertainty in the measured data, a longer 2D edge signifies more reliable information in the grasp extraction step. On the other hand, appearance of an edge in the image is relied on the distance and orientation of the object w.r.t. the camera view. Thus, depth pixel density of an object in 2D image affects the detection performance and reliability of its corresponded grasp.

### B. Robot Experiments

For the real world experiments, the approach is run in two phases, grasp planing and grasp execution. In the first phase, the proposed approach is applied to the sensed data and extracted grasping options are presented to the user by displaying the candidate pairs of contact regions. Based on the selected candidate, a 3D grasp is computed for the execution phase and the grasp strategy is performed. During all the experiments, arm manipulator, RGBD camera and the computer station are connected through a ROS network. The right arm of Baxter is fitted out with a parallel gripper. The gripper is controlled with two modes, in its "open mode" fingers distance is manually adjusted , $\epsilon_{\max} = 7cm$ based on the size of the utilized objects. During the "closed mode", fingers take either minimum distance, $\epsilon_{\min} = 2cm$ or hold a certain force value in the case of contacting. The grasp strategy is described for the end effector by taking the following steps:
Step 1) Move from an initial pose to the planned pre-grasp pose.
Step 2) Wend through a straight line from pre-grasp pose to final grasp pose with fingers in the open-mode.
Step 3) Contact the object by switching the fingers to the close-mode.
Step 4) Lift the object and move to post-grasp pose.
In the current implementation, pre-grasp and post-grasp poses have the same orientation as the final grasp pose. While, they hold a position, $20cm$ above the final grasp position. In this way, the end-effector approaches the object while holding a fixed orientation. Consequently, the fingers are prevented from colliding with the object prior to the grasp. Notice that a motion planner is utilized to find feasible trajectories for the arm joints.

We defined two scenarios to examine algorithm's overall performance, single object and multiple objects setups. In all the experiments, we assume target object is placed in the camera field of view, there exists at least one feasible grasp based on the employed gripper configuration and planned grasps are in the workspace of the robot. An attempt is considered as a *successful grasp*, if the robot could grasp the target object and hold it for a 5 sec duration after elevating. In the cases, where the user desired object does not provide a grasp choice, the algorithm acquires a new image from the sensor. If the grasp does not show up even in the second try, we consider the attempt as a failed case. In the case, where planned grasp is valid but the motion planner fails to plan or execute the trajectory, the iteration is discarded and a new query is called.

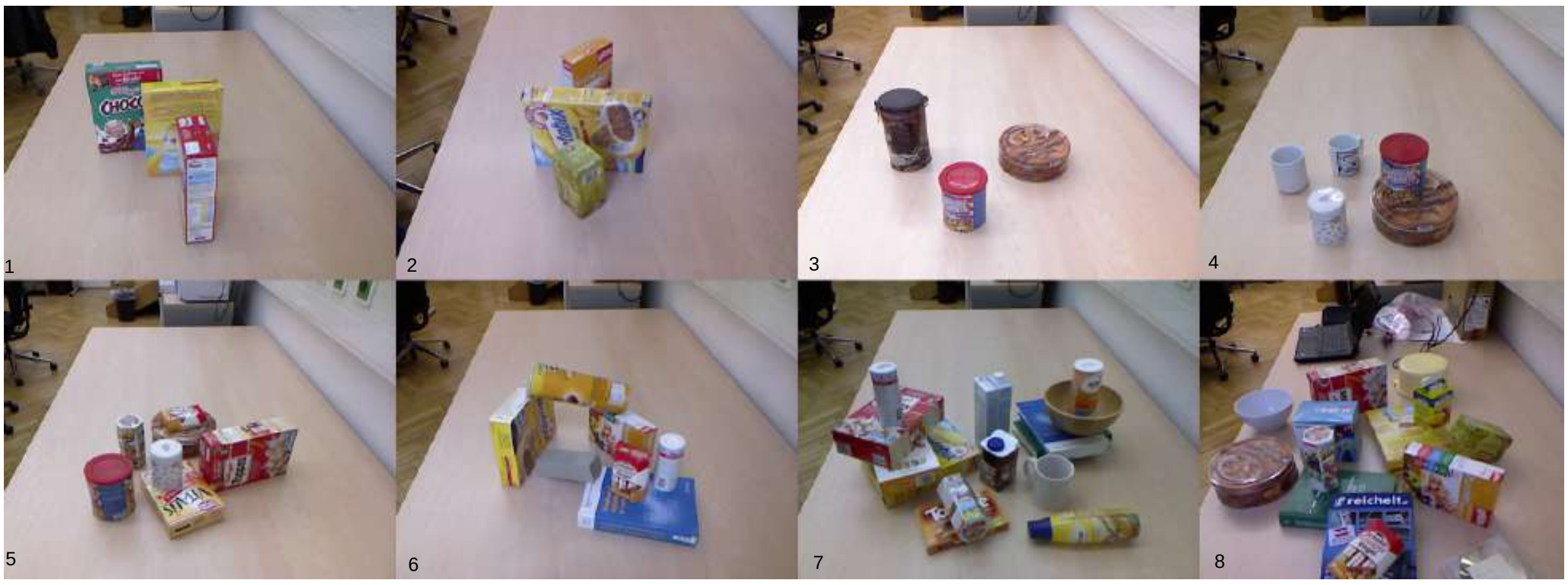

Fig. 11. Utilizede images for obtaining simulation results.

| Scene | Objects | G. Object | D. Object | G. Surface | D. Surface | G. Edge | D. Edge |
|---|---|---|---|---|---|---|---|
| No.1 | Boxes | 3 | 3 | 6 | 6 | 17 | 14 |
| No.2 | Boxs | 3 | 3 | 8 | 8 | 20 | 17 |
| No.3 | Cylinders | 3 | 3 | 6 | 5 | 12 | 10 |
| No.4 | Cylinders | 5 | 5 | 10 | 9 | 20 | 19 |
| No.5 | Mixed - low cluttered | 6 | 6 | 13 | 9 | 28 | 21 |
| No.6 | Mixed - low cluttered | 7 | 7 | 13 | 9 | 28 | 22 |
| No.7 | Mixed - high cluttered | 11 | 11 | 24 | 17 | 55 | 42 |
| No.8 | Mixed - high cluttered | 12 | 10 | 22 | 16 | 49 | 33 |
| Average detection accuracy rate | | 97% | | 81% | | 80% | |

TABLE I
SIMULATION SECTION RESULTS. COLUMNS DESCRIBE THE NUMBER OF (G)RASPABLE AND (D)ETECTED OBJECTS, SURFACE SEGMENTS AND EDGE FOR 8 DIFFERENT SCENE. THE LAST ROW INDICATES AVERAGE ACCURACY RATES OF DETECTION IN OBJECT LEVEL, SURFACE-LEVEL AND EDGE-LEVEL.

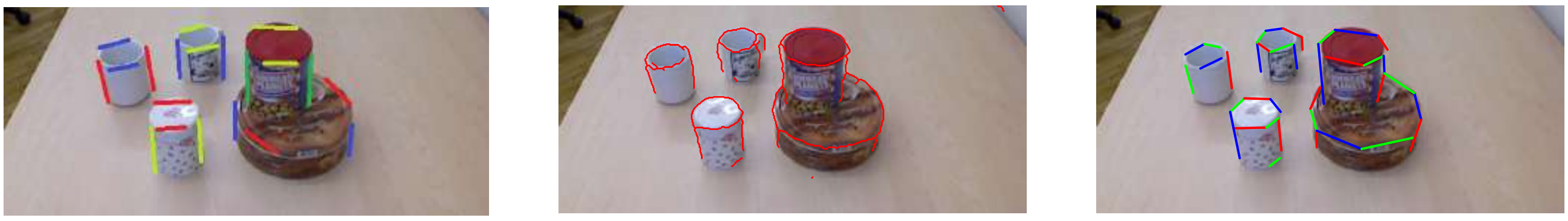

Fig. 12. Reference and detected edges for scene No.4 in the simulation-based results. Note that assigned colours are only used to distinguish the line segments visually. (a) reference graspable edges: each edge is manually marked by a line segment (b) detected graspable edges: marked points are detected by algorithm as graspable edges (c) detected line segments: each detected edge is represented by a number of line segments.

In single object experiments, objects are in an isolated arrangement on a table in front of the robot. Four iterations are performed, covering different positions and orientations for each object. The grasp is planned by the algorithm provided in the previous section followed by robot carrying out the execution strategy to approach the object. Prior to conducting each experiment, finger relative position of the Baxter gripper are set to be wide enough for the open mode and narrow enough for the closed mode. Figure (13) displays all the objects were used in the experiments and Table (VI-B) shows the obtained results in single object experiment.

According to the provided rates, 90% of the robot attempts were successful for the entire set where 11 objects were grasped successfully in all 4 iterations, 4 objects were failed to be grasped successfully in 1 out of 4 iterations and one

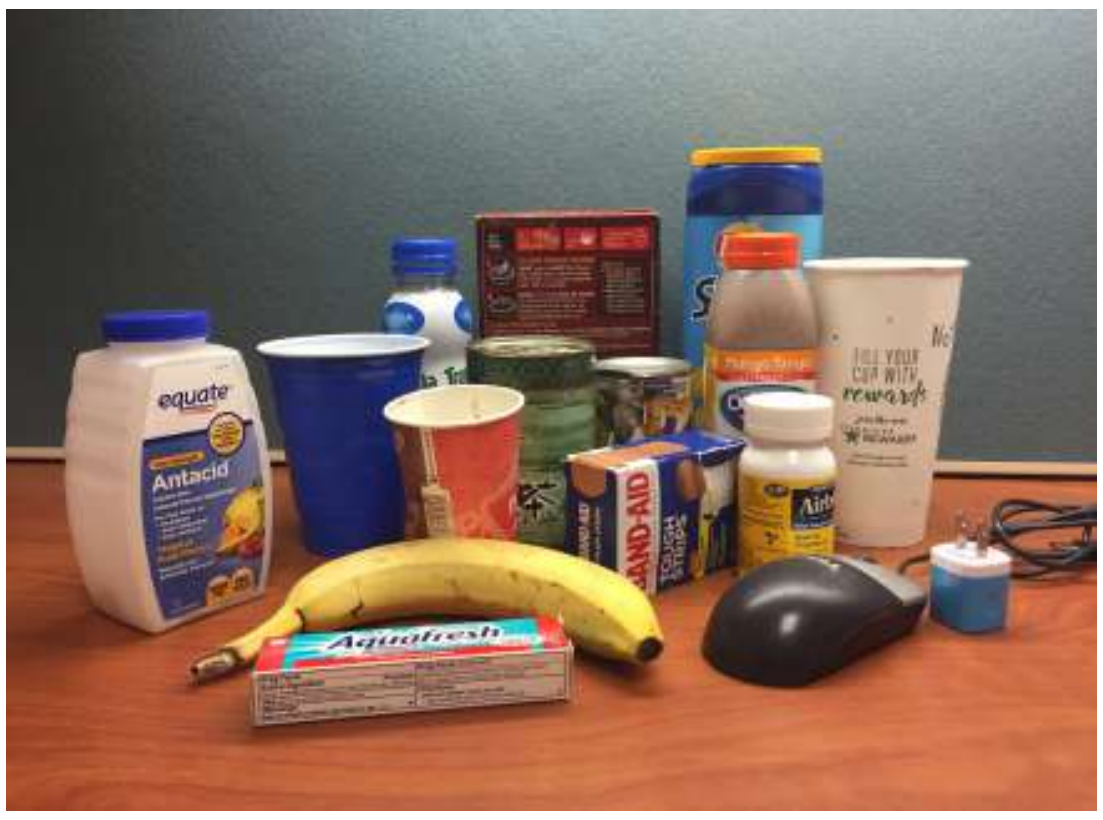

Fig. 13. The entire set of objects used through real world experiments (16 objects).

| Object | % Succ. | Object | % Succ. |
|---|---|---|---|
| Toothpaste Box | 100 | L Box | 100 |
| S Blue Box | 75 | L Paper Cup | 100 |
| Banana | 100 | L Plastic Cup | 100 |
| S Paper Cup | 75 | Green Cylinder | 100 |
| Apple Charger | 75 | L Pill Container | 100 |
| Tropicana Bottle | 100 | Chips Container | 75 |
| S Pill Container | 100 | Smoothie Bottle | 100 |
| Mouse | 50 | Fruit Can | 100 |
| Average: 90.62 % | | | |

TABLE II
SINGLE OBJECT EXPERIMENT RESULTS. FOUR ATTEMPTS FOR EACH OBJECT ARE PERFORMED. "L" INDICATES THE LARGE SIZE AND "S" INDICATES SMALL SIZE OBJECTS.

object (mouse) had 2 successful and 2 unsuccessful attempts.

In the unsuccessful attempts, the inappropriate orientation of the gripper during approaching moment is observed as the main reason of failure (4 out of 6) preventing the fingers from forming force closure on the desired contact regions. Basically, this evokes performance of plane extraction from the detected contact regions. Observation during the experiments, illuminate high sensitivity of the plane retrieval step to existence of unreliable data in the case of curved shape objects. For instance, in grasping the toothpaste box, although estimated normal direction ($\vec{V}_R$) made a $19^o$ angel with the expected normal direction (actual normal of the surface), the object was lifted successfully. While, a $9^o$ degree error resulted in failure of grasping the mouse. Impact of force closure uncertainties on the mouse case is also noticeable. For the other 2 unsuccessful attempts in the single object experiment, inaccurate positioning of the gripper was the main reason for the failure. For grasping the apple charger, gripper could not contact the planned regions, due to noisy values retrieved from low number of pixels on the object edges.

Multi object experiments are conducted to demonstrate the algorithm overall performance in a more complex environment. In each scene, a variety of objects are placed on the table and the robot approaches the object of interest in each attempt. Measuring the reliability and quality of candidate grasps are not in the scope of this paper. Hence, the order of grasping objects is manually determined such that:
i) the objects which are not blocked by other objects in the view, are in the first order.
ii) the objects which lifting them, result in scattering the other objects are in the last order.
Therefore, the user specifies select one of the candidate grasps and robot attempts the target object unless there is no feasible grasps in the image. This experiment includes 6 different scenes, two scenes with box shaped objects, two scenes with cylinder shaped objects and two scenes with variety of shapes. Table (VI-B) indicates the obtained results of multi object experiment and figure (14) demonstrates the setups of three of the scenes.

| Scene No. | Objects | Grasped Objects | Total attempts |
|---|---|---|---|
| 1 | Boxes | 4 out of 4 | 4 |
| 2 | Boxes | 5 out of 5 | 5 |
| 3 | Cylinders | 4 out of 5 | 6 |
| 4 | Cylinders | 5 out of 5 | 6 |
| 5 | Mix. | 5 out of 6 | 8 |
| 6 | Mix. | 5 out of 8 | 8 |

TABLE III
MULTI OBJECT EXPERIMENT RESULTS. THE SUCCESS RATE IMPLIES NUMBER OF OBJECTS GRASPED SUCCESSFULLY OUT OF TOTAL NUMBER OF OBJECTS IN THE SCENE.

Based on the obtained results, the proposed approach yields in 100% successful rate for box shaped objects, 90% for curved shapes and 72% for very cluttered scenes with mixed objects. Figure (15) indicates a sequence of images during the grasp execution for the scene No.3 in the multi object experiment. The robotic arm attempted to grasp the existing cylinder shaped objects located on the table. In the first two attempts, orange and blue bottles were successfully grasped, lifted and removed from the scene. Although in the third attempt the gripper contacted the paste can and elevated it, but the object was dropped caused by lack of sufficient friction between the fingertips and object surface. Then, the arm approached the remained objects (Green cylinder and large paper cup) and grasped them successfully. In the last attempt, another grasp was planned for the paste can by capturing a new image. This attempts was also failed because of inaccurate estimated pose. Finally, the experiment was finished with one extra attempt and successful rate of 4 out of 5.

### C. Discussion

According to the implemented approach, we discuss the performance of the approach and failure reasons in three levels, 2D contact region detection, 3D grasp extraction and execution. In the detection phase, the output is a pair of 2D line segment. Formation of false positive pairs and loss of false negative pairs are caused by the following reasons:
i) inefficiency of edge detection.
ii) incorrect identification of edge type feature (DD/CD).

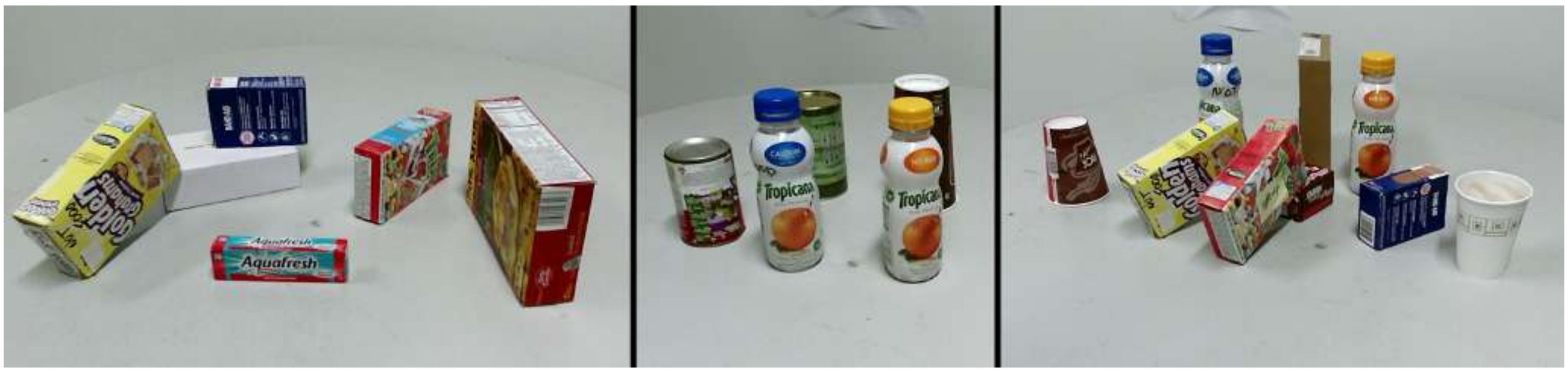

Fig. 14. Multi-object experiments scenes including variety of objects. (a) Scene No.2 (b) Scene No.3 (c) Scene No.6

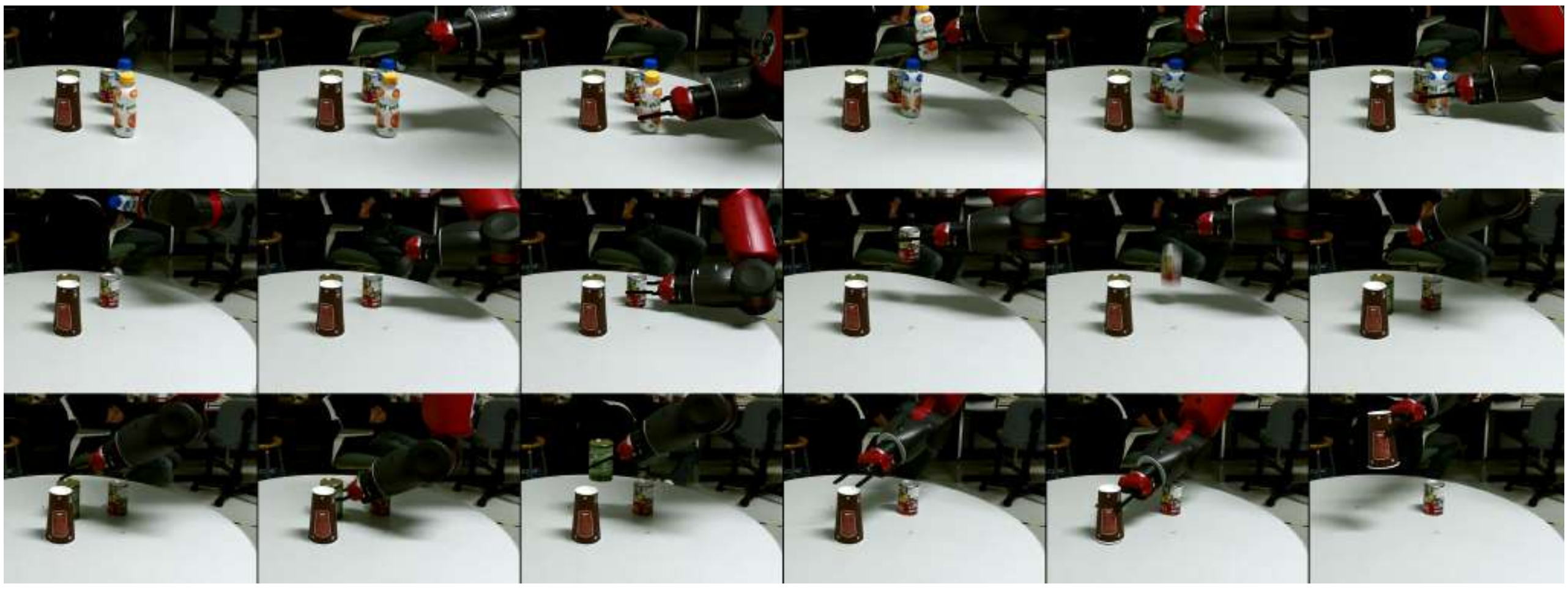

Fig. 15. A Sequence of snapshots from the robot arm while approaching to grasp the objects in a cluttered scene.

iii) incorrect identification of wrench direction feature $(\pm, +, -, 0)$.
Generally, the above reasons ensue from the measurement noise and appearance of artifacts in the data. However, objective modification based on specific datasets can yield in the performance improvement. Note that if we perform detections on a synthetic dataset without adding noise, these reasons do not affect the output.

Since the user selects a desired pair, false positive output of detection phase, do not impact the grasp attempt in the conducted experiments. As a matter of fact, in 3D grasp extraction step, the approach provides grasp parameters $(P_G, \vec{V}_R, \vec{V}_C)$ based on a true positive pair of contact regions. In overall, the grasp parameter estimation errors source from the following reasons:
i) inaccurate DD edge pixel placement (foreground /background).
ii) unreliable data for low pixel density objects.
iii) noise in the captured data.
Since we derive contact regions instead of contact points, deviation of $P_G$ in certain directions is negligible unless the finger collides an undesired surface while approaching the object. Wideness of the gripping area with respect to the target surface determines limits for this deviation. Further, error in estimation of $\vec{V}_R$ also yields in exerting force to improper regions and consequently results in unsuccessful grasp. Sensitivity of a grasp to this parameter, depends on the surface geometry and fingers kinematic. Compliance fingers show high flexibility to the estimated plane error, while firm wide fingertips do not tolerate the error. Uncertainties and assumptions regarding the friction coefficient, robot calibration and camera calibration errors are among the factors impacting the performance of execution step. Future work will focus on three directions: 1) extracting more geometric features from the available data to control the uncertainties, 2) employing efficient techniques to reduce the noise effects and 3) equipping the approach with a process to evaluate the grasp quality and reliability.

## VII. Conclusions

We propose an approach to grasp novel objects in an unstructured scene. Our algorithm estimates reliable regions on the contours (formed by a set of depth edges) to contact the object based on geometric features extracted from a captured single view depth map. The proposed algorithm leads to a force-closure grasp. Real world experiments demonstrates the ability proposed method to successfully grasp a variety of objects in shape, size and colors.